\documentclass[letterpaper]{article} % DO NOT CHANGE THIS
\usepackage{aaai25}  % DO NOT CHANGE THIS
\usepackage{times}  % DO NOT CHANGE THIS
\usepackage{helvet}  % DO NOT CHANGE THIS
\usepackage{courier}  % DO NOT CHANGE THIS
\usepackage[hyphens]{url}  % DO NOT CHANGE THIS
\usepackage{graphicx} % DO NOT CHANGE THIS
\usepackage{natbib}  % DO NOT CHANGE THIS AND DO NOT ADD ANY OPTIONS TO IT
\usepackage{caption} % DO NOT CHANGE THIS AND DO NOT ADD ANY OPTIONS TO IT
\usepackage{algorithm}
\usepackage{algorithmic}
\usepackage{amsmath}
\usepackage{xcolor}

\usepackage{newfloat}
\usepackage{listings}
\DeclareCaptionStyle{ruled}{labelfont=normalfont,labelsep=colon,strut=off} % DO NOT CHANGE THIS
\floatstyle{ruled}
\newfloat{listing}{tb}{lst}{}
\floatname{listing}{Listing}
\title{\textit{\huge NeSyCoCo}: A \textit{\huge Ne}uro-\textit{\huge Sy}mbolic \textit{\huge Co}ncept \textit{\huge Co}mposer for Compositional Generalization}
\author{
    Danial Kamali,
    Elham J. Barezi,
    Parisa Kordjamshidi
}
\affiliations{
    Michigan State University\\
    \{kamalida, jebalbar, kordjams\}@msu.edu
    
}

\usepackage{bibentry}
\begin{document}

\maketitle

\begin{abstract}
Compositional generalization is crucial for artificial intelligence agents to solve complex vision-language reasoning tasks. Neuro-symbolic approaches have demonstrated promise in capturing compositional structures, but they face critical challenges: (a) reliance on predefined predicates for symbolic representations that limit adaptability, (b) difficulty in extracting predicates from raw data, and (c) using non-differentiable operations for combining primitive concepts. To address these issues, we propose NeSyCoCo, a neuro-symbolic framework that leverages large language models (LLMs) to generate symbolic representations and map them to differentiable neural computations. NeSyCoCo introduces three innovations: (a) augmenting natural language inputs with dependency structures to enhance the alignment with symbolic representations, (b) employing distributed word representations to link diverse, linguistically motivated logical predicates to neural modules, and (c) using the soft composition of normalized predicate scores to align symbolic and differentiable reasoning. Our framework achieves state-of-the-art results on the ReaSCAN and CLEVR-CoGenT compositional generalization benchmarks and demonstrates robust performance with novel concepts in the CLEVR-SYN benchmark.
\end{abstract}
\begin{links}
\link{Code}{https://github.com/HLR/NeSyCoCo}
\end{links}

\section{Introduction}

\begin{figure*}[h]
    \centering
    \includegraphics[width=\linewidth]{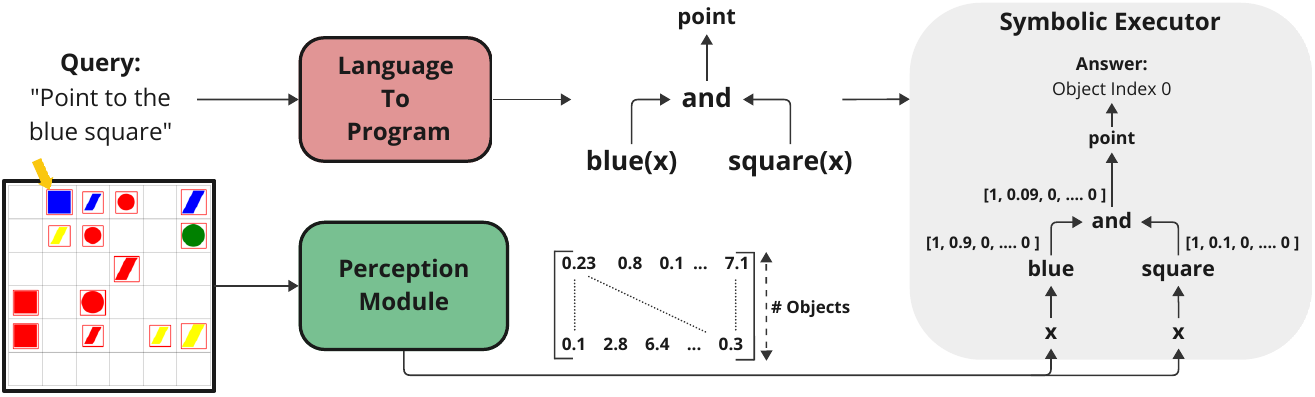}
    \caption{
The overall framework of NeSyCoCo. The language-to-program module generates a logical program based on the input query. Predicates, such as \texttt{blue}, serve as symbolic representations connected to neural modules that process representations of visual elements. These modules produce scores indicating the applicability of the concept to these elements. Differentiable soft compositional operations are then applied to the scores, executing the program and generating the answer to the query.      
}
\label{fig:overal-arch}
\end{figure*}

Compositional generalization refers to the ability of an intelligent agent to extend its understanding from previously seen components to more complex problems. Our research focuses on vision-language reasoning, an area where compositional generalization plays a crucial role. 

While humans can easily extrapolate their understanding of primitive concepts to more complex problems, current state-of-the-art models frequently encounter difficulties in this area, particularly in reasoning about the composition of entity properties or relationships~\cite{partee1984compositionality}.

Neuro-symbolic methods have demonstrated great potential in addressing compositional structures~\cite{zhu2022generalization}. However, existing approaches face several challenges. They require a symbolic representation of the domain, typically involving a set of predefined predicates, which limits the flexibility and coverage of linguistic lexical variety in concepts appearing in language and vision modalities. Moreover, obtaining the domain predicates from the raw modalities at an appropriate level of abstraction is challenging. Additionally, integrating symbolic and neural models requires differentiable operations for composing primitive concepts, which still poses a challenge.

Recent advancements in artificial intelligence have aimed to emulate human reasoning through various models and techniques. Leveraging large language models for visual reasoning has shown notable promise, with methods such as VisProg~\cite{Gupta2022VisProg} and ViperGPT~\cite{suris2023vipergpt} demonstrating potential by utilizing LLMs to generate reasoning programs for visual tasks. These models decompose complex linguistic inputs into logical steps, similar to human reasoning. However, a significant limitation is their reliance on a limited set of predefined predicates, which restricts their flexibility and ability to generalize to new scenarios. LEFT~\cite{hsu2024s} addresses the need for predefined predicates in symbolic reasoning by leveraging LLMs to extract trainable and lexically-motivated predicates. Despite this improvement, LEFT struggles with lexical variety in linguistic expressions and handling non-canonical or unseen concepts.

Research indicates that end-to-end vision-language models often struggle with compositional generalization~\cite{zhu2022generalization, yun2022vision}, frequently failing to generalize beyond their training examples. This limitation highlights the need for approaches that dynamically adapt to new predicates and queries without being constrained by predefined rules~\cite{zhu2022generalization}. In this work, we propose NeSyCoCo, a novel visual reasoning method designed to address above-mentioned limitations. Following ~\citet{hsu2024s}, our method leverages the power of LLMs, using their natural language vocabulary as a source of predicates and symbols, thus alleviating the need for manual engineering of domain predicates. We augment the linguistic inputs with their syntactic structure to improve the semantic alignment of the symbolic representations generated by LLM. In addition, we use distributed representations of concepts in language as predicate representations and connect these predicates to neural modules to deal with a lexical variety of concepts. We normalize these predicates' outputs, making it easier to compose them with soft operations that align better with task semantics. Soft composition in our framework involves the normalization of predicate scores and the application of composition functions specifically designed to operate effectively on these normalized values, ensuring that all predicates contribute to the composition equally and effectively.

NeSyCoCo excels in reasoning and grounding across multiple problems without relying on a limited set of predicates. It achieves state-of-the-art results on the ReaSCAN compositional generalization and CLEVR-Puzzle benchmarks, while maintaining high accuracy when encountering new and similar concepts in our newly created CLEVR-SYN evaluation benchmark. In summary, our contributions are:
\begin{enumerate}
    \item \textbf{Improved Symbolic Reasoning Engine}: We present an advanced symbolic reasoner that enhances the composition of primitives and interpretability, facilitating better understanding and analysis of the reasoning process.
    \item \textbf{Handling Language Variety}: Our approach handles the language variety of predicates by utilizing the predicate's distributed linguistic representation.
    \item \textbf{Introducing an Enhanced Prompting Method}: We propose an improved prompting technique for translating linguistic input to the symbolic programs using additional syntactical information as context, leading to more semantically aligned programs.
\end{enumerate}

\section{Related Works}

Our framework integrates neural networks, logical reasoning, and large language models to achieve compositional generalization in vision-language reasoning. Therefore, we focus on the four topics below.

\subsection{Compositional Generalization in Vision-Language Reasoning}
Compositional generalization is a crucial aspect of AI systems, enabling them to handle more complex compositions in vision-language tasks. Recent studies have examined the generalization capabilities of various neural network architectures using specialized evaluation tasks \cite{Hupkes2019, ontanon, csordas}. Benchmarks such as CLEVR \cite{johnson2017clevr}, gSCAN \cite{ruis2020benchmark}, and ReaSCAN \cite{Wu-ReaSCAN} have been developed to assess these capabilities in vision language models. Recent works have introduced advanced transformer-based architectures~\cite{kamali-kordjamshidi-2023-syntax, Sikarwar2022, jiang-bansal-2021-inducing, qiu-etal-2021-systematic}, special neural architectures~\cite{kuo2021compositionalnetworksenablesystematic, gao2020systematic, hsu2023disco}, meta learning~\cite{xu-etal-2023-metarevision, Xu2023MetaCR} and soft prompting~\cite{xu2024gipcol}, to address compositional generalization. A recent survey overviews compositional learning approaches from theoretical and experimental perspectives, offering insights into the field~\cite{sinha2024a}.

Unlike previous end-to-end methods, we employ a neuro-symbolic approach. Recent work has shown that neuro-symbolic methods perform slightly worse than end-to-end models on in-domain problems but better in generalization \cite{zhu2022generalization}. NeSyCoCo can generalize to new similar concepts and utilizes soft composition for more primitive concepts.

\subsection{Neuro-Symbolic Vision-Language Reasoning}

Neuro-symbolic approaches have demonstrated strong vision-language reasoning capabilities by combining symbolic reasoning with neural networks through modular designs. For instance, Neuro-symbolic VQA~\cite{yi2018neural} and the Neuro-Symbolic Concept Learner (NSCL) \cite{Mao2019NeuroSymbolic} have advanced visual reasoning by utilizing symbolic program execution and reducing the need for dense supervision. Despite their successes, these approaches often rely on predefined domain-specific languages and manually implemented programs, which limits their flexibility. The recent model, LEFT~\cite{hsu2024s}, utilizes LLMs to generate symbolic representations for linguistic queries, partially alleviating this problem. However, this method struggles with handling the lexical variety in predicate language and fails when faced with novel concepts, demonstrating limited generalization capability. NeSyCoCo addresses these limitations by employing LLMs and normalized predicate outputs for soft composition. This enables more interpretable and flexible compositional generalization. Unlike previous approaches, NeSyCoCo does not depend on a limited set of predefined symbols and can adapt to various predicates, enhancing its ability to generalize across different tasks.

\subsection{LLMs for Formal Representation}
Leveraging large language models to decompose tasks into sequences of API calls has gained attention in recent research~\cite{cheng2022binding, beurer2023prompting, zelikman2023parsel,10.1007/978-3-031-71170-1_25}. These methods typically focus on the natural language domain, limiting their capacity to ground concepts in visual or other modalities. While LLMs can reason about object categories inferred from language, they cannot recognize objects in a scene or generate robotic actions. Some approaches~\cite{Gupta2022VisProg, suris2023vipergpt} utilize LLMs to generate programs to execute on images but rely on predefined modules without additional training. Similar to LEFT~\cite{hsu2024s}, our approach overcomes these limitations by using LLMs to obtain formal representations and leveraging their natural language vocabulary for predicates and symbols, enhancing flexibility and coverage in grounding concepts. Additionally, we employ dependency parsing as an additional context for LLM to improve symbolic program generation.

\subsection{General Vision-Language Models}
Vision-language models (VLMs) have shown success in multimodal environments. These models integrate vision and language modalities to perform tasks such as image captioning~\cite{xiao2023florence}, visual question answering~\cite{Qwen2VL,liu2024llavanext}, and navigation~\cite{zhang2023vln,zhang2024vision}. Despite their success, end-to-end VLMs often struggle with generalization in novel and complex tasks across different domains \cite{zhu2022generalization, yun2022vision}. This limitation arises from their end-to-end nature, which makes it challenging to handle the diverse and intricate relationships between vision and language components. Our approach marks competitive performance with VLMs when faced with new, complex problems.

\section{Methodology}
This work addresses the challenge of compositional generalization in vision-language reasoning tasks. Our approach involves a bi-modal input system, where the inputs consist of a natural language query and an image that provides context. The objective is to answer the query given the image context accurately.

\begin{table*}[h!]
    \centering
    \setlength{\tabcolsep}{1mm}
    \small
    \begin{tabular}
    {|l|l|l|c|c|}
        \hline
        \textbf{Function} & \textbf{Logical Form} & \textbf{Description} & \multicolumn{2}{c|}{\textbf{Differentiable Implementations}} \\
        \cline{4-5}
        & & & \textbf{LEFT} \small \cite{hsu2024s} & \textbf{NeSyCoCo} \\
        \hline
        exists($\alpha_x$) & \(\exists(\alpha_x)\) & Existential quantification & \(\max(\alpha_x)\) & \(\max(\alpha_x)\) \\
        \hline
        forall($\alpha_x$) & \(\forall(\alpha_x)\) & Universal quantification & \(\min(\alpha_x)\) & \(\min(\alpha_x)\) \\
        \hline
        and($\alpha_x, \alpha_y, \alpha_z$) & \( \alpha_x \land \alpha_y \land \alpha_z\) & Logical conjunction & min(\(\alpha_x, \alpha_y, \alpha_z\)) & \(\alpha_x \odot \alpha_y \odot \alpha_z\) \\
        \hline
        and($\alpha_x, \beta_{xy}$) & \(\alpha_x \land \beta_{xy}\) & Logical conjunction & \(\sum_{y}(\alpha_x \odot \beta_{xy})\) & \(\max_{y}(\alpha_x \odot \beta_{xy})\) \\
        \hline
        not($\alpha_x$) & \(\neg(\alpha_x)\) & Logical negation & \(- \alpha_x\) & \(1 - \alpha_x\) \\
        \hline
        iota(var, $\alpha_x$) & \(\iota(\text{var}, \alpha_x)\) & Variable assignment & softmax(\(\alpha_x\)) & \( \frac{\alpha_x - \min(\alpha_x)}{\max(\alpha_x) - \min(\alpha_x)} \) \\
        \hline
        count($\alpha_x$) & \(\text{count}(\alpha_x)\) & Counting elements & \(\sum\sigma(\alpha_x)\) & \(\sum\alpha_x\) \\
        \hline
        equal($s_1$, $s_2$) & \( s_1 == s_2 \) & Scalar equality & \( \sigma \left( \frac{\tau \cdot (\gamma - |s_1 - s_2| )}{\gamma} \right) \)  & \( \sigma \left( \frac{\tau \cdot (\gamma - |s_1 - s_2| )}{\gamma}  \right) \) \\
        \hline
        greater\_than($s_1$, $s_2$) & \(s_1 > s_2\) & Scalar inequality & \(\sigma (\tau \cdot (s_1 - s_2 - 1 + \gamma))\) & \(\sigma (\tau \cdot (s_1 - s_2 - 1 + \gamma))\) \\
        \hline
    \end{tabular}
    \caption{Mathematical Expressions: Logical Forms, Descriptions, and Differentiable Implementations.}
    \label{tab:expressions}
\end{table*}

We introduce the Neuro-Symbolic Concept Composer (NeSyCoCo), a unified neuro-symbolic framework designed to interpret natural language queries by decomposing them into differentiable symbolic functions. These functions are then combined using soft composition techniques to generate accurate responses. As illustrated in Figure \ref{fig:overal-arch}, the Neuro-Symbolic Concept Composer comprises three key components:

\begin{enumerate}
    \item \textbf{Natural Language to Program}: Converts natural language queries into symbolic programs, forming the basis for reasoning.
    \item \textbf{Perception Module}: Extracts domain-specific features, such as objects or relational features, from the input data.
    \item \textbf{Differentiable Neuro-Symbolic Reasoning Executor}: Executes the symbolic programs, composes the relevant concepts, and generates the final answer to the query.
\end{enumerate}

Our approach utilizes the LEFT implementation~\cite{hsu2024s}, serving as the foundation for our framework. We build on this framework by introducing three key improvements: 1) leveraging dependency parsing to achieve more accurate symbolic representations of language in natural language to program, 2) reducing reliance on predefined symbolic predicates by utilizing linguistically motivated distributed representations in neuro-symbolic reasoner, and 3) refining the compositional operations for the soft execution of symbolic programs in neuro-symbolic reasoner.

\begin{figure}[h]
    \centering
    \includegraphics[width=\linewidth]{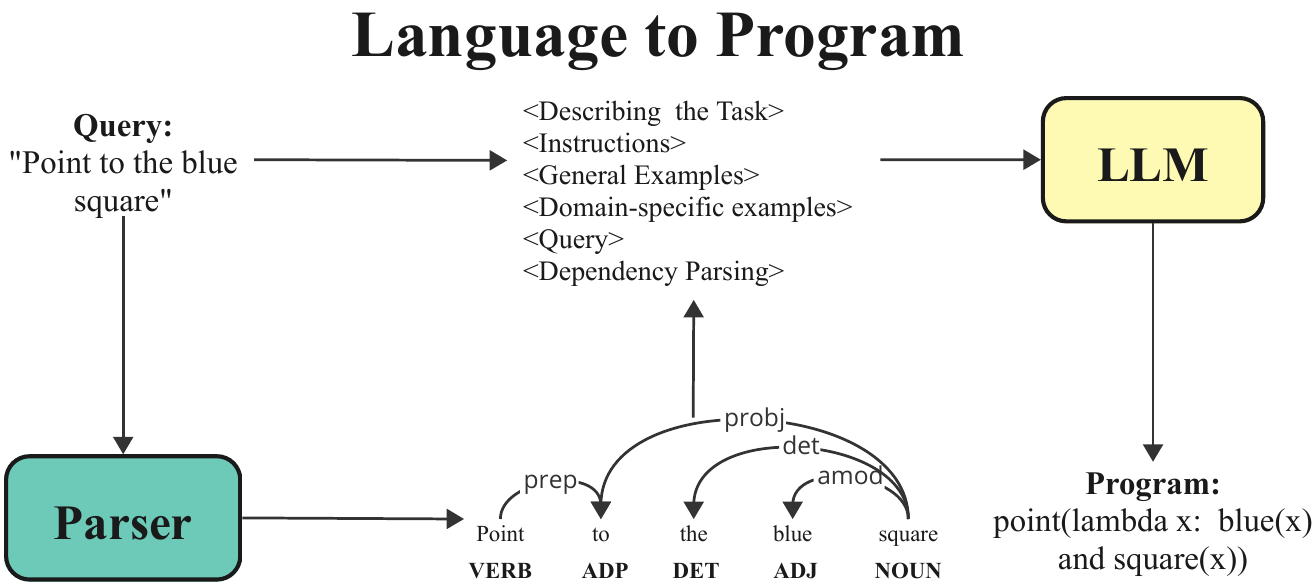}
    \caption{Language to program conversion procedure.}
    \label{fig:language-to-code}
\end{figure}

\subsection{Natural Language To Program: Exploiting Dependency Parsing}

The Natural Language To Program component converts natural language queries into symbolic representations, as illustrated in Figure~\ref{fig:language-to-code}. We use an LLM to generate a program based on the linguistic query. Following LEFT~\cite{hsu2024s}, the prompt provided to the backbone model includes syntactical instructions and general examples, supplemented with simple domain-specific examples when necessary to guide the symbolic program generation.

Previous work did not elaborate on improving the natural language to program conversion and simply used LLMs for program generation or assumed it is given~\cite{hsu2024s}. However, understanding complex nested and compositional expressions can be challenging for large language models, particularly in capturing structural dependencies. To address this, and inspired by previous work ~\cite{johnson2017inferringexecutingprogramsvisual,kamali-kordjamshidi-2023-syntax}, we incorporate dependency parsing to aid in symbolic program generation. We use the dependency parsing of the query and represent it as a sequence. 
We enable the model to exploit the dependency structure of the queries to generate better semantically aligned programs.

For instance, consider the query \textit{point to the blue square}, as shown in Figure \ref{fig:language-to-code}. In this process, the query is first parsed, resulting in a dependency structure such as \texttt{[square, pobj, to, ADP, [the, blue]]}. This parsed dependency information is then concatenated with the original query and provided as additional context to the language model. The language model then translates this input into the symbolic program \texttt{point(lambda x: blue(x) and square(x))}, as illustrated in Figure \ref{fig:language-to-code}.

\subsection{Perception Module}
The Perception Module allows the model to integrate information across multiple modalities. This module extracts features or representations from a secondary modality, such as 2D images. While the perception module can be adapted for various domains, we focus on image perception in this work to support vision-language reasoning. For images, we extract entity-centric representations using Mask RCNN~\cite{DBLP:journals/corr/HeGDG17}, ResNet~\cite{he2016deep}, and PreciseRoIPooling \cite{jiang2018acquisitionlocalizationconfidenceaccurate}. Specifically, given an image with $N$ bounding boxes, this process returns a tensor representation $\mathbf{E}_o^{N \times d_o}$, where $d_o$ is the dimensionality of the visual feature vectors, with each row corresponding to features extracted from an individual bounding box. Additionally, it yields a tensor $\mathbf{E}_r^{N \times N \times d_r}$ to represent relational features for each pair of bounding boxes in the image.

\begin{figure}[h!]
    \centering
    \includegraphics[width=0.98\linewidth]{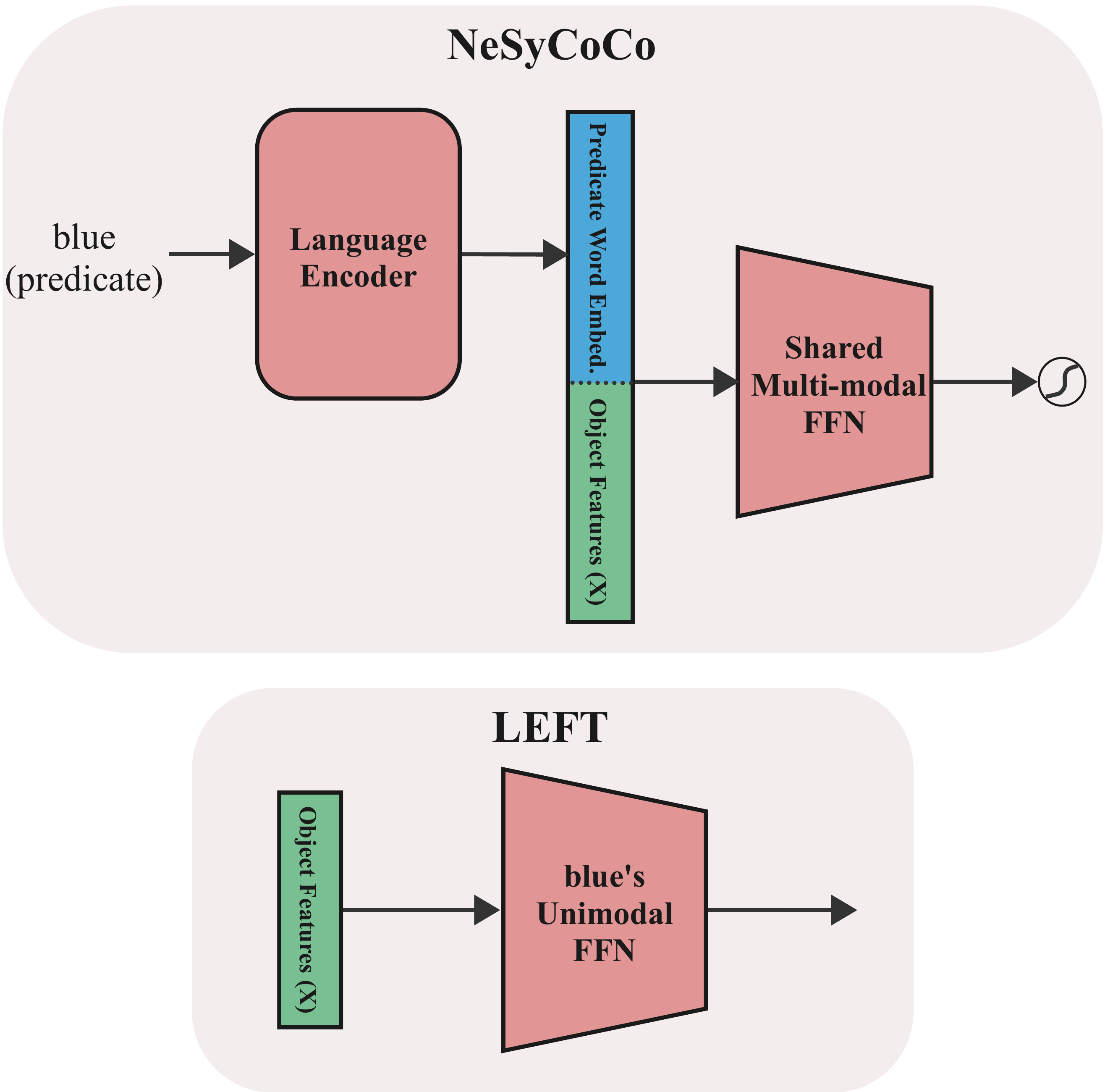}
    \caption{Differentiable predicate function in NeSyCoCo (shared FFN for all predicates) compared to LEFT (predicate-specific FFNs) calculating the score for \texttt{blue}.}
    \label{fig:predicate_function}
\end{figure}

\subsection{Differentiable Neuro-Symbolic Reasoning Module}
This module executes the programs once the natural language queries are converted into symbolic programs and the visual representations are extracted. These programs consist of two types of functions: domain predicates and first-order logic operations.

\subsubsection{Domain Predicates: Exploiting Linguistically-motivated Distributed Representations of Symbolic Predicates} 
Domain predicates are neural functions that assign values to objects or relations based on specific traits. For single-object (unary) predicates, functions such as \texttt{big(x)} or \texttt{red(x)} generate a score $\alpha_x$, indicating the extent to which object $x$ exhibits the predicate’s characteristic. For multi-object (binary and ternary) predicates, such as \texttt{same\_row(x, y)}, the function returns a score $\beta_{xy}$, reflecting the relationship between objects $x$ and $y$ according to the predicate. 
In contrast with previous works, to address the problem of linguistic lexical variety, we utilize distributed linguistic representation. Therefore, in NeSyCoCo, each domain predicate, regardless of its object-arity, such as \texttt{inside(., .)} or \texttt{red(.)} is represented with a vector embedding from an off-the-shelf language encoder resulting in $\mathbf{E}_w^{1 \times d_w}$. As shown in Figure \ref{fig:predicate_function}, at the domain predicate level, the predicate's embedding is concatenated with the perception module's representation and passed to a shared multilayer Feed Forward Network (FFN) to generate a score for the combination of linguistic predicate and the perceived visual representation. For parallelization, the $\mathbf{E}_w^{1 \times d_w}$ is repeated to calculate the scores for multiple objects simultaneously.

\[
\begin{aligned}
\alpha &= \sigma\left(FNN(\mathbf{E}_p^{N \times d_{o}} \| \mathbf{E}_w^{N \times d_w})\right) \\
\beta &= \sigma\left(FNN(\mathbf{E}_r^{N \times N \times d_{r}} \| \mathbf{E}_w^{N \times N \times d_w})\right)
\end{aligned}
\]
This approach alleviates the need for canonical predicates and helps learn the parameters of linguistic predicates with similar semantics.

\subsubsection{First Order Logic Executor: Utilizing Soft Composition}
\label{sec:method-fol}
NeSyCoCo programs execute recursively starting with an expression such as \texttt{point(lambda x: expr)} or \texttt{count(lambda x: expr)}, where a variable is defined, and a sub-expression is embedded. This sub-expression is processed recursively through tensor functions. We employ soft operations rather than relying on the hard, discrete operations typical of traditional symbolic reasoning. This approach allows us to backpropagate errors throughout the pipeline, enabling effective training of primitive predicates. 

The LEFT approach to symbolic execution has two major issues. First, it uses scalar values, which complicates the composition of scores across different scales. In contrast, NeSyCoCo utilizes a sigmoid activation function at the predicate level to normalize the scores. This technique controls predicate scores, minimizes error ranges, and reduces the risk of propagating failures, enhancing model robustness. Moreover, it allows us to use multiplication instead of the \texttt{min} function for composition, ensuring that all predicates contribute to the composition equally. It also allows for simultaneous training of multiple concepts through backpropagation, which is impossible when using the \textit{min} function.

Second, using softmax in the \texttt{iota} variable assignment function presents another challenge. When the model encounters multiple matching objects for an expression, softmax can either reduce the output score if the objects receive similar scores or exaggerate differences in scores, making reasoning more difficult. For example, if two objects have similar scores for a predicate, applying softmax can inflate the difference between their scores, distorting the reasoning process by assigning inflated scores. To alleviate this limitation, we utilize linear normalization at the \texttt{iota} function. 

To ensure compatibility, we updated other composition functions to align with this normalization strategy, as shown in Table~\ref{tab:expressions}. By addressing these issues, our approach significantly improves composition and interoperability, leading to superior performance in vision-language tasks, particularly in compositional generalization.

\section{Experiments}
We evaluate our method across three key aspects: compositional generalization, vision-language reasoning, and handling linguistic variety. We present our experiments on ReaSCAN~\cite{Wu-ReaSCAN} and CLEVR-CoGenT~\cite{johnson2017clevr} for compositional generalization. In the context of visual reasoning, we discuss our experiments and findings using the CLEVR dataset and its extensions. Finally, to assess how our neuro-symbolic methods handle linguistic variety, we introduce a new benchmark called CLEVR-SYN.
\subsection{Experimental Setting}

Our implementation is based on the PyTorch deep learning library~\cite{Paszke2019PyTorchAI}, with the SpaCy toolkit~\cite{spacy} used for extracting dependency parsing of natural language queries. We employed the LLaMA-3.1 70B model~\cite{dubey2024llama3herdmodels} with 4-bit quantization as our primary language model, selected for its open-source availability and performance parity with GPT-3.5, allowing us to maintain transparency and adaptability in our experiments. The experiments' vector embeddings for predicates and other linguistic components were derived from GloVe~\cite{glove} language encoder. More details about the experimental settings can be found in Appendix \ref{appendix:b}.

\subsection{Compositional Generalization}

\begin{table*}[h!]
    \centering
{
\setlength{\tabcolsep}{1mm}
\small

\begin{tabular}{|l|c|c|c|c|c|c|c|c|}
\hline
& \textbf{A1 (\%)} & \textbf{A2 (\%)} & \textbf{A3 (\%)} & \textbf{B1 (\%)} & \textbf{B2 (\%)} & \textbf{C1 (\%)} & \textbf{C2 (\%)} & \textbf{Avg (\%)} \\
\hline

 GroCoT  & 99.4 &85.7 &95.4 &90.4 &83.5 &70.2 &27.7 &78.5 \\

Syntax Guided Transformer &  \textbf{99.6} & \textbf{97.3}  & \textbf{99.6} & 95.4& 90.1 & 92.5 & 21.7 &85.2\\
\hline
    
NeSyCoCo {\small (Full)}  & 99.1 {\small$\pm$} {\small 0.14} & 94.1 {\small$\pm$} {\small 0.02} & 98.5 {\small$\pm$} {\small 0.37} & \textbf{98.9} {\small$\pm$} {\small 0.05} & \textbf{98.7} {\small$\pm$} {\small 0.08} & \textbf{96.8} {\small$\pm$} {\small 0.12} & \textbf{95.9} {\small$\pm$} {\small 0.05} & \textbf{97.5} {\small$\pm$} {\small 0.05} \\

{\small w/o Embedding} & 99.5 {\small$\pm$} {\small 0.05} & 91.7 {\small$\pm$} {\small 0.41} & 99.1 {\small$\pm$} {\small 0.24} & 97.6 {\small$\pm$} {\small 0.18} & 97.7 {\small$\pm$} {\small 0.16} & 94.5 {\small$\pm$} {\small 0.16} & 95.9 {\small$\pm$} {\small 0.06} & 96.5 {\small$\pm$} {\small 0.07} \\

{\small w/o Emb. w/o Soft Reasoner}   & 97.8 {\small$\pm$} {\small 0.35} & 80.3 {\small$\pm$} {\small 1.13} & 95.5 {\small$\pm$} {\small 0.82} & 93.0 {\small$\pm$} {\small 0.25} & 95.8 {\small$\pm$} {\small 0.95} & 92.2 {\small$\pm$} {\small 0.82} & 92.9 {\small$\pm$} {\small 0.20} & 92.5 {\small$\pm$} {\small 0.23} \\

LEFT$^\dagger$& 97.8 {\small$\pm$} {\small 0.35} & 80.3 {\small$\pm$} {\small 1.13} & 95.5 {\small$\pm$} {\small 0.81} & 93.0 {\small$\pm$} {\small 0.25} & 95.8 {\small$\pm$} {\small 0.95} & 92.2 {\small$\pm$} {\small 0.82} & 91.5 {\small$\pm$} {\small 0.16} & 92.3 {\small$\pm$} {\small 0.14} \\

% ------
\hline

    \end{tabular}
    \caption{The accuracy of our proposed model on the ReaSCAN test split grounding task compared to neuro-symbolic and end-to-end methods such as GroCoT~\cite{Sikarwar2022} and Syntax Guided Transformer~\cite{kamali-kordjamshidi-2023-syntax}.$^\dagger$ The results on LEFT are reported without dependency parsing in the context. {\small w/o Emb. w/o Soft Reasoner} shows the results of LEFT with improved prompting. The reported results are the average accuracy and standard deviation of three runs.}
    \label{tab:reascan-result}
}
\end{table*}

\begin{table}[h]
{
\small
\centering
\begin{tabular}{|l|l|l|l|}
\hline
\textbf{Method} &  \textbf{Split A (\%)}  & \textbf{Split B (\%)} \\ \hline

MDETR{~\cite{kamath2021mdetr}} & \small  \textbf{99.7} & \small  76.2 \\ \hline
LEFT{~\cite{hsu2024s}} & \small  99.5   & \small  76.2 \\ \hline
NeSyCoCo & \small 99.6 {$\pm$ 0.08}  & \small  \textbf{78.8}{ $\pm$ 0.15}  \\ \hline

% NeSyCoCo & \small 99.58{$\pm$0.08}  & \small  \textbf{78.79}{$\pm$0.15}  \\ \hline

\end{tabular}
\caption{Accuracy on the CLEVR-CoGenT benchmark reported on the average of three runs.}
\label{tab:cogent}
}
\end{table}

To demonstrate our method's compositional generalization capability, we evaluated NeSyCoCo using two commonly used benchmarks, CLEVR-CoGenT and ReaSCAN. 
\subsubsection{CLEVR CoGenT Benchmark}

The CLEVR Compositional Generalization Task (CoGenT) extends the original CLEVR dataset to test model generalization to novel unseen combinations of visual attributes. This task has two splits, each with distinct attribute distributions. In test split A, cubes are restricted to gray, blue, brown, or yellow, while cylinders are limited to red, green, purple, or cyan. Split B swaps these color sets between cubes and cylinders. Spheres in both splits can appear in any color. Models are trained on biased distributions and tested on unseen attribute combinations, challenging them to develop compositional representations rather than relying on memorization.

We evaluate the model performance using the exact match accuracy of the generated response. As shown in Table \ref{tab:cogent}, our model achieved state-of-the-art results in generalization, outperforming both LEFT and MDETR~\cite{kamath2021mdetr} methods. This outcome is consistent with recent work~\cite{yun2022vision}, suggesting that neuro-symbolic approaches surpass end-to-end methods in generalization while showing slightly worse in-domain performance. Our model’s performance on CLEVR-CoGenT indicates its strong ability to address the compositional challenges in this benchmark, demonstrating robust generalization.

\subsubsection{ReaSCAN Benchmark} 
To further examine our model's compositional generalization capabilities, we evaluated it on the ReaSCAN \cite{Wu-ReaSCAN} dataset, which is specifically designed to test compositional generalization in grounded language understanding. ReaSCAN consists of instructions for an agent to perform tasks within a 2D environment, requiring an understanding of spatial relationships and object properties. The goal is to assess how well vision-language models generalize from familiar linguistic inputs to novel combinations of learned concepts. This dataset is crucial for testing model's ability to interpret linguistic commands, offering insights into its capacity for compositional and grounded language processing. ReaSCAN includes seven compositional test splits with specific held-out combinations compared to the training data:

\begin{itemize}
    \item A1: \texttt{yellow square} referred with color and shape.
    \item A2: \texttt{red square} referred anywhere in the command.
    \item A3: \texttt{small cylinder} referred to by size and shape.
    \item B1: Co-occurrences of a \texttt{small red circle} and a \texttt{large blue square}.
    \item B2: Co-occurrences of \texttt{same size as} and \texttt{inside of} relationships.
    \item C1: Three relative clause commands.
    \item C2: Two relative clause using \texttt{that is} instead of \texttt{and}.
\end{itemize}
Recent research identifies grounding as the key challenge in the ReaSCAN dataset, and accurate grounding enables perfect navigation step generation~\cite{Sikarwar2022}. Thus, we concentrated on grounding in ReaSCAN, leaving navigation for future work. Therefore, we use the accuracy of the \textit{object localization} for evaluation.

As shown in Table~\ref{tab:reascan-result}, our model outperformed the baseline on this compositional generalization benchmark and surpassed previous non-symbolic methods across B and C test splits while showing competitive results on A split. In addition, our analysis of the wrong cases revealed that our method has difficulty handling size-related concepts. In ReaSCAN, size is a contextual and relative concept, while we handle size using unary object-level functions. Hence, our method struggles to interpret these concepts accurately.

We performed an ablation study to show the significance of each proposed component of our approach. As shown in Table~\ref{tab:reascan-result}, our three technical components contribute to obtaining the SOTA performance. Among them, the soft symbolic reasoner has the most substantial impact on compositional generalization, as evidenced by a paired t-test (\( p = 0.0026 \)), confirming that the observed improvements are statistically significant. In addition, employing word embedding for predicate representation marks a high performance by handling a variety of predicates. As a matter of fact, obtaining an accurate program is a crucial step to achieving a precise reasoning model, which was provided by an improved prompting method that shows improvement, mostly in complex cases such as C1 and C2 test splits.

\begin{figure}[h]
    \centering
   \includegraphics[width=1\linewidth]{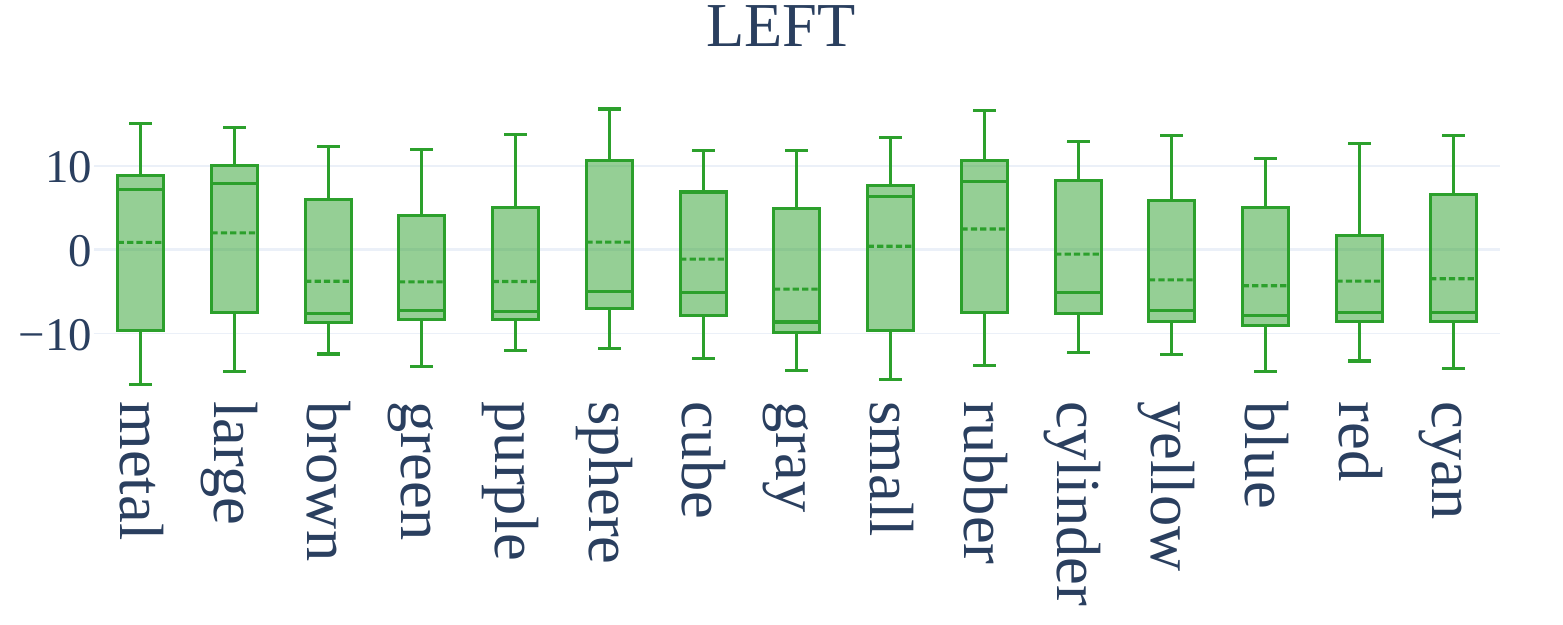}
   \includegraphics[width=1\linewidth]{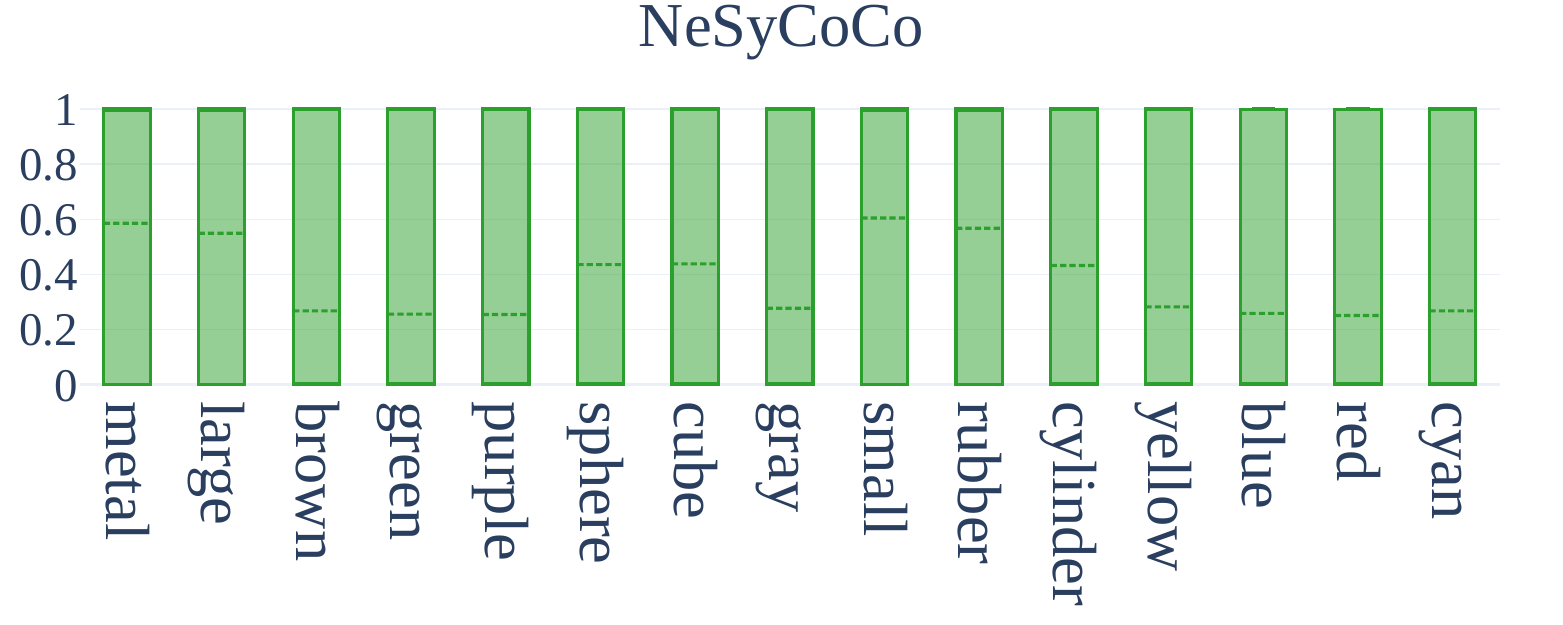}
   
    \caption{Boxplot comparing concept scores of LEFT and NeSyCoCo on 10k CLEVR validation samples. Dotted and solid lines represent the mean and median, respectively.}
    \label{fig:left-clevr-concept-scores}
\end{figure}

As mentioned in Section \ref{sec:method-fol}, LEFT~\cite{hsu2024s} uses raw unbounded logits as concept scores. We analyzed the scale of concept scores given by the LEFT predicate function on the CLEVR dataset, as illustrated in Figure \ref{fig:left-clevr-concept-scores}; the score ranges for different concepts can vary significantly. This becomes problematic when combining concepts like \texttt{red} and \texttt{rubber} using a \textit{min} function. In such cases, the score for \texttt{red} is often undervalued, leading to a biased composition that fails to reflect the true relationship between the concepts. As shown in the ablation study in Table~\ref{tab:reascan-result}, NeSyCoCo outperforms previous work on compositional generalization largely due to its use of soft composition functions for combining primitives during symbolic program execution, which shows the positive effect of our modifications.

\subsection{Vision-Language Reasoning}

To evaluate our method's vision-language reasoning capability, we evaluated our model's performance on the CLEVR dataset and some of its extensions.

\subsubsection{CLEVR and Extensions} 
The CLEVR dataset~\cite{johnson2017clevr} is a benchmark for vision-language reasoning, featuring synthetic images of diverse objects and associated questions assessing tasks like counting, attribute comparison, and spatial understanding. CLEVR's structured design effectively tests models' compositional learning and reasoning capabilities using the accuracy of generated answers. As shown in Table~\ref{tab:clevr}, our model outperforms previous neuro-symbolic methods and demonstrates competitive performance to end-to-end models showing its advanced vision-language reasoning.

\begin{table}[h!]
\centering
{
\setlength{\tabcolsep}{1mm}
\small
\begin{tabular}{|l|l|l|l|}
\hline
\textbf{Method} & \textbf{Type} & \textbf{Accuracy} (\%)  \\ \hline

MDETR{~\cite{kamath2021mdetr}} & \small End-to-End &  \small \textbf{99.7}\\ \hline

% NSCL{~\cite{Mao2019NeuroSymbolic}}$^{\dagger}$ & \small Neuro-Symbolic & \small 98.74   \\ \hline
NSCL{~\cite{Mao2019NeuroSymbolic}}$^{\dagger}$ & \small Neuro-Symbolic & \small 98.7   \\ \hline

% LEFT {~\cite{hsu2024s}} & \small Neuro-Symbolic & \small 99.62 \\ \hline

LEFT {~\cite{hsu2024s}} & \small Neuro-Symbolic & \small 99.6 \\ \hline

NeSyCoCo & \small Neuro-Symbolic  & \small \textbf{99.7 }{$\pm$ 0.02}  \\ \hline

\end{tabular}

\caption{Accuracy on the validation set of the CLEVR dataset reported on the average of three runs. $\dagger$ indicates the use of pre-defined predicates}
\label{tab:clevr}
}
\end{table}

We also evaluate our method on CLEVR-based extensions introduced in~\citet{hsu2024s}, including CLEVR-Ref for referring expressions, CLEVR-Puzzle for multi-step reasoning, and CLEVR-RPM for abstract reasoning. As shown in Table \ref{tab:clevr-transfer}, our model achieves 100\% accuracy on CLEVR-Ref and CLEVR-RPM, and 95\% on CLEVR-Puzzle, outperforming all previous models. Human evaluations indicate that the five errors in the CLEVR-Puzzle were due to incorrect annotations. Otherwise, the models would also obtain 100\% accuracy on this test setting.

\begin{table}[h!]
\centering
{
\setlength{\tabcolsep}{1mm}
\small 

\begin{tabular}{|l|l|l|l|}
\hline
\textbf{Model} & \textbf{Ref} &  \textbf{Puzzles} &  \textbf{RPM} \\ \hline
NeSyCoCo {+ GT programs} & \textbf{100}\% & \textbf{95}\% & \textbf{100}\% \\ \hline
LEFT {+ GT programs} & 100\% & 92\% & 100\% \\ \hline
LEFT {+ LLM programs} & 94\% & 75\% & 87\% \\ \hline
LLaVA NeXT{\small\cite{li2023llava}} & N/A & 45\% & 52\% \\ \hline
OpenFlamingo{\small\cite{awadalla2023openflamingo}} & N/A  & 57\% & 52\% \\ \hline
ViperGPT{~\cite{suris2023vipergpt}}  & 8\% & 34\% & 4\% \\ \hline
VisProg{~\cite{Gupta2022VisProg}} & 35\% & 27\% & 51\% \\ \hline

\end{tabular}
\caption{Accuracy on CLEVR extension tasks.} 
\label{tab:clevr-transfer}
}
\end{table}

% \subsubsection{Data Efficacy}
% \subsubsection{CLEVR}
\subsection{Linguistic Lexical Variety}

\subsubsection{CLVER-SYN}
Since we are using a frozen vector-based representation of predicates, our model should be able to deal with new and similar concepts when faced with new and similar concepts. To showcase this capability, we created a new benchmark based on the CLEVR dataset validation set called the CLEVR synonym (CLEVR-SYN) benchmark for neuro-symbolic methods. This benchmark aims to evaluate the performance of a neuro-symbolic method trained with original CLEVR programs when faced with new concepts. This benchmark consists of three test splits using all of the samples in the CLEVR validation split. In the test splits of this dataset, concepts in programs have been replaced with unseen but similar primitive concepts. Table \ref{tab:syn-pearson} shows the concepts and their substitutes. The splits of this dataset are easy, medium, and hard tests. In the easy test, only one concept has changed in the program. In the medium test, a maximum of three concepts have been replaced. In the hard test, all concepts in the list are replaced. We replaced these concepts using regular expressions search in the programs given by the CLEVR dataset. 

We evaluated our model using these three different splits. As shown in Table \ref{tan:syn-results}, our model maintains high accuracy on the CLEVR-SYN benchmark and outperforms previous work in zero-shot concept generalization. The previous approaches falter mainly due to their inability to handle new but semantically similar predicates effectively. They will either fail due to the lack of trained predicates or resort to random initialization for new predicates.

\begin{table}[h!]
    \centering
    {
\setlength{\tabcolsep}{1mm}
\small
\begin{tabular}{|c|c|c|c|}
    \hline
    \textbf{Predicate} & \textbf{Similar Concept} & \textbf{$\rho$} &  \textbf{p-value} \\ \hline
cube& box    & 0.06      & 0.44      \\ \hline
sphere       & ball   & -0.02     & 0.49      \\ \hline
large        & huge   & 0.97      & 0.01      \\ \hline
small        & little & 0.91      & 0.04      \\ \hline
metal        & metallic        & 0.69      & 0.15      \\ \hline
rubber       & elastic& 0.68      & 0.18      \\ \hline
red          & burgundy        & 0.15      & 0.47      \\ \hline
blue         & azure & 0.55      & 0.34      \\ \hline
brown        & chocolate       & 0.20      & 0.44      \\ \hline
yellow       & mustard& 0.05      & 0.51      \\ \hline
left         & left\_of        & 1.00      & 0.00      \\ \hline
front        & front\_of       & 0.98      & 0.01      \\ \hline
same\_color  & matching\_color & 0.86      & 0.06      \\ \hline
same\_material & identical\_material & 0.75 & 0.11      \\ \hline
same\_shape  & congruent\_shape & 0.66 & 0.17      \\ \hline

    \end{tabular}
    \caption{CLEVR-SYN substitutions predicates and their score's Pearson correlation score CLEVR validation set.}
    \label{tab:syn-pearson}
    }
\end{table}

\begin{table}[h!]
\centering
{
\setlength{\tabcolsep}{1mm}
\small
\begin{tabular}{|l|l|l|l|l|}
\hline
\textbf{Method} & \textbf{Easy} (\%) & \textbf{Medium} (\%) & \textbf{Hard} (\%) \\ \hline
% TbD & - & 71.0   & 52.8 & 45.6 \\ \hline
LEFT & 81.9 {$\pm$ 0.19}  & 64.8 {$\pm$ 0.26} & 49.5 {$\pm$ 0.64} \\  \hline 
% LEFT & 81.7{$\pm$0.19}  & 64.7{$\pm$0.27  } & 56.2 \\  \hline 

% NeSyCoCo  & 92.12{ $\pm$0.26}  & 81.19{ $\pm$0.58} & 73.42 { $\pm$0.66} \\  \hline 
NeSyCoCo  & \textbf{92.1 }{$\pm$ 0.26}  & \textbf{81.2} {$\pm$ 0.58} & \textbf{73.4} {$\pm$ 0.66} \\  \hline 
\end{tabular}
}
\caption{Accuracy on the CLEVR synonym benchmark reported on the average of three runs.}
\label{tan:syn-results}
\end{table}

\subsubsection{Analysis} We further analyzed the synonym dataset to evaluate our method's performance on individual substitutions by measuring the correlation between replaced concept scores and the original predicate's scores using the Pearson correlation ($\rho$)~\cite{Benesty2009Pearson} metric and p-value to show the significance of the correlation. As shown in Table \ref{tab:syn-pearson}, our model effectively captures nuanced relationships across various attributes, as evidenced by significant correlations between semantically similar concepts. Notably, terms like \texttt{large/huge} and \texttt{small/little} exhibit strong positive correlations with high p-values with their corresponding learned concepts. The model also shows strong generalization for multi-token predicates like \texttt{same material/identical material}, crucial for handling non-canonical predicates in a neuro-symbolic system.

However, while our method performs well in 9 out of 15 cases (correlation higher than 0.6), it encounters difficulties with certain predicates, such as \texttt{ball} vs \texttt{sphere} or \texttt{brown} compared to \texttt{chocolate}. These challenges likely arise from the nuanced semantic or contextual differences between these terms, which frozen embeddings do not capture fully. For example, \texttt{brown} and \texttt{chocolate} may overlap semantically as two similar colors but often differ in contextual usage and sensory perception.

\begin{figure}[h!]
    \centering
    \includegraphics[width=\linewidth]{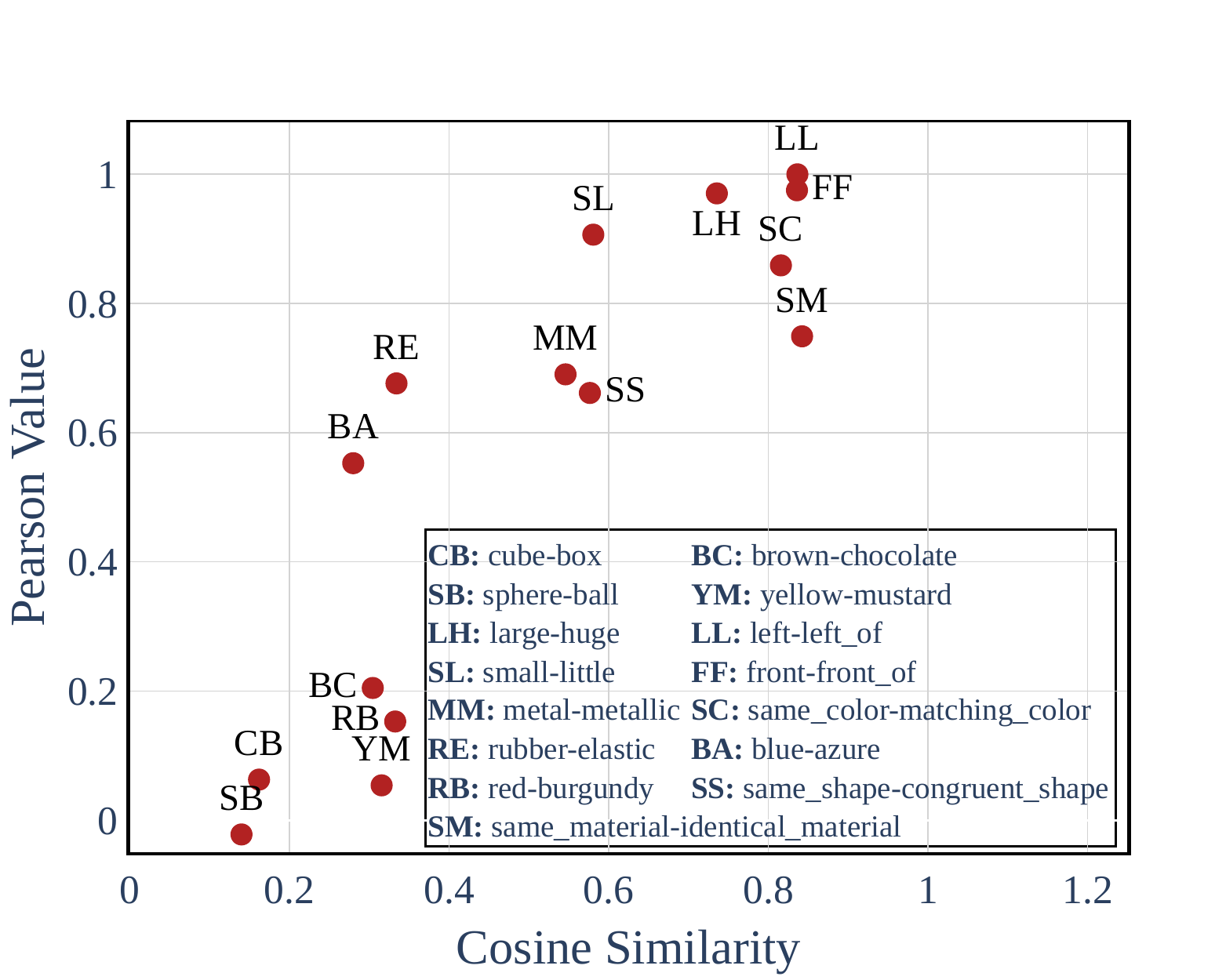}
    \caption{Relationship between cosine similarity of word embeddings and correlation of their predicate scores.}
    \label{fig:cosine_pearson_labels_adjusted}
\end{figure}

To investigate this further, we analyzed the relationship between the cosine similarity of predicates’ embeddings and the correlations in their scores. As shown in Figure \ref{fig:cosine_pearson_labels_adjusted}, there is a strong positive correlation between embedding similarity and predicate scores, suggesting that our model effectively generalizes when embeddings reflect semantic closeness (cosine similarity higher than 0.4). This finding underscores the importance of high-quality embedding representations for neuro-symbolic models, particularly for predicates with intricate or context-sensitive meanings. Addressing these limitations could enhance the model’s generalization ability across a broader range of predicates. More evaluation on different encoders can be found in Appendix \ref{appendix:a}.  

These results support our argument that our neuro-symbolic method can generalize to new, unseen concepts, especially in domains where linguistic relationships are reflected in distributed representations. This ability to generalize is crucial for developing AI systems capable of adapting to novel situations, highlighting the potential of neuro-symbolic approaches in achieving robust and flexible AI.

\section{Conclusion}
In this paper, we introduced NeSyCoCo to address fundamental challenges in the existing neuro-symbolic frameworks. We utilized the soft composition of normalized predicate functions to compose primitive concepts more effectively, thereby enhancing the model's reasoning capabilities and compositional generalization. In addition, by integrating word embeddings of the symbolic predicates, NeSyCoCo effectively addresses the challenge of linguistic variability, enabling zero-shot generalization to novel but semantically related concepts. Furthermore, using syntactical parsing as additional context for the large language model during program generation enhances the precision and accuracy of the generated symbolic programs. These contributions address challenges in bridging symbolic reasoning and neural network-based approaches. As a result, our approach demonstrated state-of-the-art performance on challenging benchmarks such as ReaSCAN and CLVER-CoGenT. One interesting future direction is to integrate the NeSyCoCo approach in the generic neuro-symbolic framework DomiKnowS~\cite{guo2020inference, faghihi2021domiknows, gluecons} to make the underlying neural models programmable instead of using a fixed neural architecture for predicates.

\appendix

\section{Choice of Language Encoder}
\label{appendix:a}
For the language encoder selection, we tested RoBERTa~\cite{liu2019robertarobustlyoptimizedbert}, Spacy~\cite{spacy}, GloVe 6B-300D~\cite{glove}, and one-hot encoding. Our experiments demonstrated that all the models performed well on the original CLEVR validation set. Notably, GloVe exhibited strong generalization on the CLEVR-SYN dataset, as presented in Table \ref{tab:syn-encoder}.

\begin{table}[h!]
{\setlength{\tabcolsep}{1mm}
\small

    \centering
    \begin{tabular}{|l|c|c|}
    \hline
        \textbf{Language Encoder} & \textbf{CLEVR}  & \textbf{CLEVR-SYN Easy}  \\ \hline
        Spacy & 99.7\% & 80.7\% \\ \hline
        RoBERTa & 99.7\% & 84.7\% \\ \hline 
        One-hot & 99.7\% & - \\ \hline
        Glove-6B-300D & 99.7\% & 92.1\% \\ \hline
    \end{tabular}
    \caption{Accuracy of NeSyCoCo with different language encoders on the CLEVR dataset and CLEVR-SYN Easy split.}
    \label{tab:syn-encoder}
}
\end{table}

\section{Experimental Setting}
\label{appendix:b}
All experiments were conducted on Ubuntu OS with an AMD EPYC 7413 24-core CPU and an NVIDIA A6000 GPU, featuring 48GB of memory and 700GB of RAM. The code, generated data, and necessary dependencies listed in the code are available in the Github repository. The hyperparameters used in the experiments are detailed in Table \ref{tab:hyperparams}.

\begin{table*}[h!]
{\setlength{\tabcolsep}{1mm}
\small

\centering
\begin{tabular}{|l|c|c|c|}
\hline
\textbf{Hyperparameters} & \textbf{ReaSCAN} & \textbf{CLVER} & \textbf{CLVER-CoGenT} \\
\hline
Shared FNN & [1024,512, 256, 128, 1] & [1024,512, 256, 128, 1] & [1024,512, 256, 128, 1] \\ \hline
Visual Repr. Projection & 512 & 512 & 512 \\ \hline
Predicate Repr. Projection & 512 & 512 & 512 \\ \hline
Learning Rate & \{10$^{-3}$, \textbf{10$^{-4}$}, 10$^{-5}$\} & \{10$^{-2}$ ,10$^{-3}$ ,\textbf{10$^{-4}$}, 10$^{-5}$\} & \{10$^{-3}$ ,\textbf{10$^{-4}$}, 10$^{-5}$\} \\ \hline
Batch Size & 32 & 32 & 32 \\ \hline
Number of Parameters & 14.2M & 14.3M & 14.3M  \\ \hline
Epochs & 100 & 100 & 100 \\ \hline
Curriculum Learning & Yes & Yes & Yes \\ \hline
Language Encoder & Glove-6B-300D & Glove-6B-300D & Glove-6B-300D \\  \hline
Embedding Size & 300 & 300 & 300 \\ \hline
% Avg Time (1 Epoch) & 64 & 14 & 18 \\ \hline

\end{tabular}
\caption{Hyperparameters of NeSyCoCo for ReaSCAN, CLVER, and CLVER-CoGenT}
\label{tab:hyperparams}
}
\end{table*}

\section*{Ethical Statement}
While our method demonstrates considerable improvements in compositional reasoning, it is not without notable limitations. Most of our experiments were conducted on synthetic datasets, which offered a controlled environment to evaluate model performance, particularly on compositional generalization. However, these datasets may not fully encapsulate the complexity and variability of real-world scenarios, highlighting the need for future evaluations on diverse, real-world datasets to ensure broader practical applicability. Additionally, the reasoning capabilities of NeSyCoCo depend on programs generated by pre-trained language models that were not specifically optimized for this task. Although syntax errors in LLM-generated programs can be mitigated through detection and resampling, unresolved semantic errors remain a significant challenge. Lastly, while our method exhibits superior generalization ability compared to related work, its predicate generalization performance remains dependent on the choice of distributed representation, which may limit its adaptability in certain scenarios.

\section*{Acknowledgments}

This project is supported by the Office of Naval Research (ONR) grant N00014-23-1-2417. Any opinions, findings, and conclusions or recommendations expressed in this material are those of the authors and do not necessarily reflect the views of Office of Naval Research.
{
\bibliography{aaai25}
}

\end{document}